\documentclass[sigconf,nonacm]{acmart}

\AtBeginDocument{%
  \providecommand\BibTeX{{%
    \normalfont B\kern-0.5em{\scshape i\kern-0.25em b}\kern-0.8em\TeX}}}

\copyrightyear{2021}
\acmYear{2021}
\acmConference[WSDM WebTour'21]{ACM WSDM Workshop on Web Tourism (WSDM Webtour'21}{March 12, 2021}{Jerusalem, Israel}
\acmBooktitle{ACM WSDM Workshop on Web Tourism (WSDM Webtour'21), March 12, 2021, Jerusalem, Israel}
\acmPrice{}
\acmDOI{}
\acmISBN{}
\usepackage{subfigure}
\usepackage{xcolor}
\usepackage[normalem]{ulem}
\usepackage{siunitx}
\usepackage[utf8]{inputenc}
\begin{document}

\title{Modeling Multi-Destination Trips with Sketch-Based Model}

\author{Michał Daniluk}
\affiliation{%
  \institution{Synerise}
  \institution{Warsaw University of Technology}
  \country{Poland}
}
\email{michal.daniluk@synerise.com}

\author{Barbara Rychalska}
\affiliation{%
  \institution{Synerise}
  \institution{Warsaw University of Technology}
  \country{Poland}
}
\email{barbara.rychalska@synerise.com}

\author{Konrad Gołuchowski}
\affiliation{%
  \institution{Synerise}
  \country{Poland}
}
\email{konrad.goluchowski@synerise.com}

\author{Jacek Dąbrowski}
\affiliation{%
  \institution{Synerise}
  \country{Poland}
}
\email{jack.dabrowski@synerise.com}


\begin{abstract}
The recently proposed EMDE (Efficient Manifold Density Estimator) model achieves state of-the-art results in session-based recommendation. In this work we explore its application to Booking.com Data Challenge competition.  The aim of the challenge is to make the best recommendation for the next destination of a user trip, based on dataset with millions of real anonymized accommodation reservations. We achieve 2nd place in this competition. First, we use Cleora - our graph embedding method - to represent cities as a directed graph and learn their vector representation. Next, we apply EMDE to predict the next user destination based on previously visited cities and some features associated with each trip. We release the source code at: https://github.com/Synerise/booking-challenge.

\end{abstract}






\maketitle

\section{Introduction}

The goal of the challenge \cite{booking2021challenge} is to predict the final city of each trip using a dataset based on millions of real anonymized accommodation reservations. 
The released train set contains 1,166,835 unique reservations within 217,686 trips and 39,901 unique cities in 195 countries. 
A list of features is presented in Table \ref{features}.

The evaluation dataset is constructed similarly, however the city ID of the final reservation of each trip is concealed and requires a prediction. 
The test set consists of 378,667 reservations with at least 4 consecutive reservations. Predictions were made for 70,662 unique trips. The test set was drawn from the same temporal distribution as a training set. 

\textbf{Evaluation.}
The metric used for performance evaluation is precision at 4 (Precision@4). The score is understood as the average of the per-sample scores, which are either 1 if the predicted city is in top 4 predictions, or 0 otherwise. The teams were allowed to make only 2 submissions on the final test set.

\textbf{Solution.}
We frame the problem of route prediction as a session-based recommendation task. Our contributions are as follows: 
\begin{itemize}
    \item  We propose to represent cities as nodes in a directed graph, whose edges represent trips between two cities. We compute city embeddings with Cleora \cite{rychalska2021cleora}, a fast and efficient network embedding technique.
    \item We apply EMDE \cite{emde} to recommend the next destination based on representations of previous cities and additional numerical and categorical features such as the length of a trip or the type of user's device.
    \item We analyze the effectiveness and challenges of our method.
\end{itemize}

\textbf{Overall challenge results.}
Our approach takes 2nd place out of 38 in this challenge with the final Precision@4 score of 0.5780, compared to the leading score of 0.5939 and surpasses the 3th and 4th place solution scores of 0.5741 and 0.5566, respectively.

\begin{table}[h!]
\caption{Dataset statistics.}
  \label{features}
  \centering
  \small
\begin{tabular}{c|c}
\hline
Feature        & Number of unique values \\  \hline
User ID        & 200153                   \\ 
Trip ID        & 217686                 \\ 
Check-in date  & 425 (31.12.2015-27.02.2017)                \\ 
Check-out date & 425 (1.01.2016-28.02.2017)                \\ 
Affiliate ID   & 3254               \\ 
Device Class   & 3                 \\ 
Booker Country & 5                 \\ 
Hotel Country  & 195                 \\ 
City ID        & 39901                  \\
\end{tabular}
\end{table}

\begin{figure*}%
\centering
\includegraphics[width=170mm]{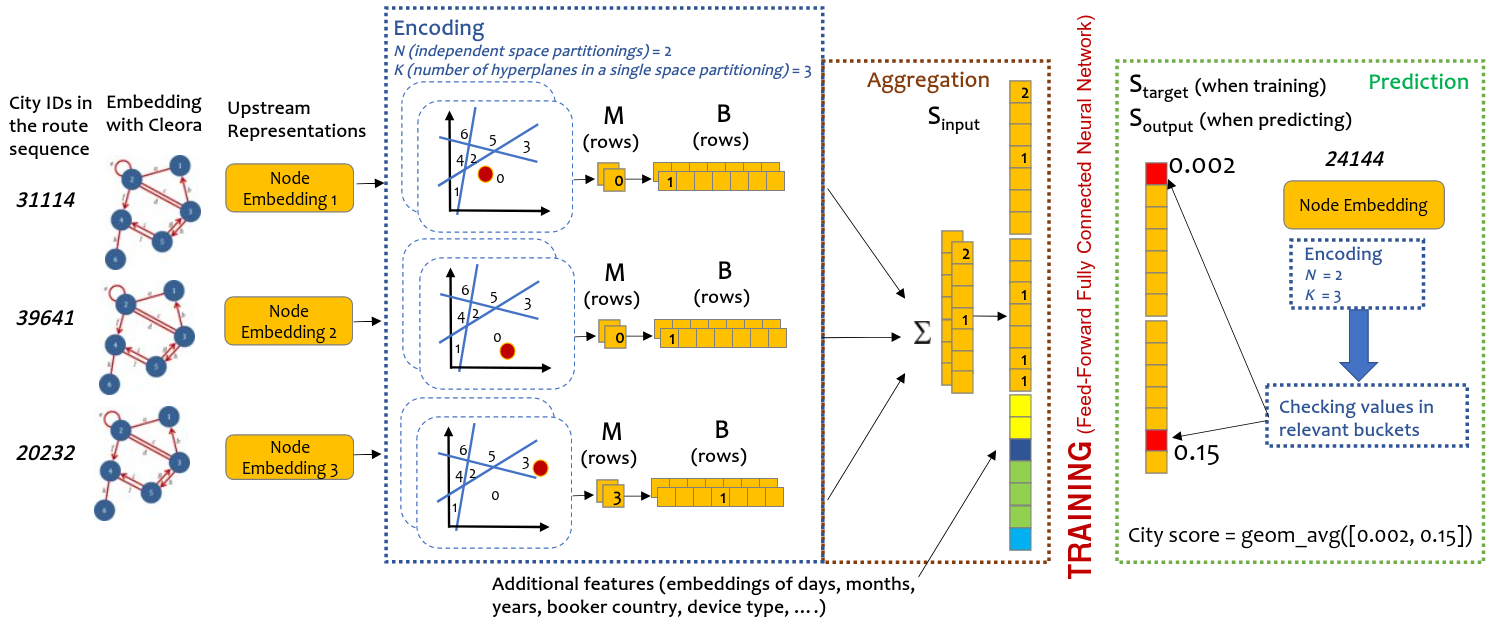}
\caption{EMDE architecture overview.}
\label{fig:architecture}
\end{figure*}

\section{Related Work}
Within the past few years, deep learning recommendation models have had difficulties in achieving good results in recommender system competitions \cite{jannach2020deep}. Instead, the winning solutions often involved gradient boosting models with substantial feature
engineering efforts. Furthermore, it has been shown that conceptually simpler techniques, e.g. based on
nearest neighbors \cite{ludewig2019performance} often outperforms neural approaches on multiple datasets.


Many deep learning session-based recommendation models consider recommendations as a sequential problem, applying recurrent networks (LSTM/GRU) \cite{hidasi2018recurrent,li2017neural,pan2020intent,ruocco2017inter}, which are known to have difficulties in learning long-term dependencies and scale poorly to growing item sets and increasing sequence lengths \cite{tallec2017unbiasing}. On the other hand, graph-based models (GNN) \cite{yu2020tagnn, wu2019session,chen2020handling} cast recommendations as a graph traversal problem. Those methods exhibit a number of specific efficiency-related problems such as \textit{neighborhood explosion} (the number of neighbors often grows exponentially when increasing node distances are considered).
Such problems demand additional remedial measures which often hurt performance \cite{bai2020ripple,chen2017stochastic,zeng2021accurate}. Yet, as the sequential aspect of recommendation is considered vital, most efforts are focused on researching even more complex neural network architectures in order to represent the ordered relations accurately.


Instead of focusing on sequential aspect, our Efficient Manifold Density Estimator (EMDE) \cite{emde} learns users' behaviors as weighted item sets: an aggregate vector (\textit{sketch}), which is a histogram-like vector representation of densities on manifolds spanned by embedding vectors. EMDE allows to compress a variable-length sequence of cities into a simple 1-D vector of constant size, which can be fed into a simple shallow feed-forward neural network. It achieves competitive results on many datasets, and has the ability to use a multi-modal vector representation of each destination.

\section{Algorithm}


The first step of our algorithm is obtaining information-rich city embeddings. We observe that the route prediction dataset can be represented as a directed graph, where each city is a node, and each directed edge represents a trip from one city to another. 
The dataset has a graph structure with a lot of revisited cities in a single trip, $62\%$ of them have at least one cycle. In addition, representing data as a graph gives the ability to have a connection between different trips.
In order to compute node embeddings we apply Cleora \cite{rychalska2021cleora}, which relies on multiple iterations of normalized, weighted averaging of each node’s neighbor embeddings, followed by normalization across dimensions. The procedure is fast and we empirically find that it produces high quality embeddings (§\ref{section:ablations}).



\subsection{EMDE}
\label{emde section}
Efficient Manifold Density Estimator (EMDE) introduced in \cite{emde} is a probability density estimator for high dimensional spaces, which is inspired by Count-Min Sketch algorithm (CMS) and locality sensitive hashing (LSH). It works by dividing the item embedding space into regions and assigning items to specific regions (buckets) based on similarity of their embedding vectors. EMDE operates on structures analogous to multidimensional histograms, called \textit{sketches}. 

An example of the full EMDE algorithm run is depicted in Figure~\ref{fig:architecture}. Full explanation and the intuitions behind the algorithm are presented in \cite{emde}. Below we give a brief summary of the algorithm steps:
\begin{enumerate}
\item \textbf{Encoding.} EMDE operates on manifolds spanned by embedding vectors. In our case, these are embeddings of cities understood as graph nodes. First, the embedding manifolds are  partitioned with a data-dependent LSH method called DLSH. As a result, LSH regions are created primarily where the data are present. Each partitioning is done with $K$ hyperplanes and is repeated $N$ independent times  (see \texttt{Encoding} section in Fig. \ref{fig:architecture}). We denote $K$ as the \textit{sketch width} and $N$ as \textit{sketch depth}. The regions of the partitioning are analogous to  hash buckets in CMS. While a single region is usually large (typically 64-256 regions are created per single partitioning), multiple independent partitionings allow to obtain a high resolution map of the manifold via intersection or ensembling. 

\item \textbf{Region Assignment.} The region IDs of each input cities are stored in matrices $M$, which are then binarized to obtain matrices $B$ (notations are consistent with \texttt{Encoding} section in Fig. \ref{fig:architecture}), which form the \textit{sketches} of individual cities.

\item \textbf{Aggregation.} The sketches of individual cities are aggregated with simple summation to obtain a sketch of the whole trip, represented by vector $S_{input}$ (see \texttt{Aggregation} section in Fig. \ref{fig:architecture})

\item \textbf{Training.} The aggregate sketches have constant size and thus can serve as input to a simple fully-connected feed-forward neural network, which is trained to output a sketch $S_{target}$ of the hidden part of the trip.

\item \textbf{Prediction.} Score prediction procedure reuses the sketches of all cities, which have been encoded in the first algorithm step. For each city, its relevant region IDs are stored in the matrices $M$. When an output sketch $S_{output}$ is produced by the network, the values contained in relevant region IDs are retrieved for each city and averaged with the geometric mean to produce a final per-city score.

\end{enumerate}



The advantages of using EMDE are numerous, especially for large datasets. The sketches are additive, and can accommodate any number of cities within a fixed-size representation.
Sketch size is independent of the number of samples and original embedding dimensions. EMDE can easily incorporate multiple modalities of input data, as well as continuous and categorical features, by simple vector  concatenation. Note that the aggregate sketches produced by EMDE inherently lose information on city ordering.

\section{Experiments}
In this section, we first describe data preprocessing (§\ref{data preparation}), then we describe our model in details, present the results, and analyze the effectiveness of our method (§\ref{model section}).

\subsection{Data Preparation}
\label{data preparation}
The goal of the competitions is to predict only the \textbf{final destination} of a each trip. 
However, in order to give the model more information, we augment the dataset to include prediction of all previous cities based on user history.
For example, we split a four-city trip into three training examples, predicting the second, third and fourth city during the trip.

\textbf{Train/Valid Split.}
We create our own validation set that imitates the hidden test set, by sampling out $70,662$ trips from the train set. The validation set, same as the test set, has only final destinations as target. In addition, trips in validation and test sets consist of cities that appear only in the training set. Our training set includes also non-final targets from the validation set. All datasets are from the same temporal distribution.


\textbf{Features.}
For each data point we compute the following categorical and continuous features: type of user's device, country from which the reservation was made,  an ID of affiliate channels, country of the hotel, the length of stay in predicted city, the number of days since beginning of a trip, the number of days till the end of a trip, number of days since last booking, number of cities in a trip, week days of check-in and check-out, month, year.

\subsection{Model}
\label{model section}
\textbf{City embeddings.}
In order to obtain city embeddings, we represent the dataset as a directed graph of city trips. Each node in the graph denotes a city, and each directed weighted edge represents the journey between them. Weight of the edge denotes the number of trips from one city to another. We construct the graph from both training and testing examples (excluding the final missing cities).
In order to embeded cities we use Cleora \cite{rychalska2021cleora} with iteration number $i=1$ and $i=3$ and embedding length of 1024. As a result, each city is represented by two embeddings which are the input to the EMDE model.

\textbf{EMDE Configuration.}
We encode all embeddings with sketches of depth $N=40$ and width $K=128$. Each embedding configuration (Cleora embeddings computed with  $i=1$ and $i=3$) is encoded separately, representing two complementary modalities of data. We observe that adding random sketch codes (not based on LSH) for each item improves the model performance, allowing the model to separate very similar cities to differentiate their popularity.

\textbf{Model.}
We train a three-layer residual feed forward neural network with 3000 neurons in each hidden layer, with leaky ReLU activations and batch normalization. The input of the network consists of:
\begin{enumerate}
    \item Three width-wise L2-normalized, concatenated sketches: \textit{first city sketch} representing the first city in a trip, \textit{prev city sketch} representing the previous city in a trip, and \textit{all cities sketch} containing all other cities. We use the \textit{first city sketch} to facilitate training, because in about $15\%$ of training examples, the final city is the same as the first city.
    In order to create a representation of a  user's behaviour in the \textit{all cities sketch}, we aggregate the sketches of cities with a simple summation (as is normally done in EMDE), multiplying them with constant decay which reduces the influence of cities which are older in time.
    \item Normalized numerical features such as number of days since beginning of a trip or number of unique cities visited so far.
    \item Categorical features that are represented by the PyTorch \texttt{nn.Embedding} layer such as the day of the week of check-in, month, year or country from which the reservation was made. The size of the embedding depends on the number of unique feature values. We set it to $120$ for previous hotel country and affiliate channel features. For other features, the size of the embedding is $20$. 
    \item A binary flag indicating if the target is the final destination (always true in case of the test set).
\end{enumerate}
The output of the model is a sketch that represents our target city. The procedure of producing the \textit{all cities sketch} is shown in Figure \ref{fig:architecture}, also showing the addition of numerical and categorical features by simple concatenation.

\textbf{Training.}
We train our model on a single nVidia GeForce RTX 2080 Ti 11GB RAM GPU card. Training takes circa 45 minutes on this configuration.
We use AdamW optimizer \cite{loshchilov2017decoupled} with first momentum coefficient of 0.9 and second momentum coefficient of 0.999\footnote{Standard configuration recommended by \cite{kingma2014adam}} with an initial learning rate of $0.0005$, weight decay of $0.01$ and a mini-batch size of 128 for optimization. 

Since the distribution of final destinations is different than distribution of non-final cities, we train the model in two stages: 1) using non-final target destinations, and 2) fine-tuning the model on the examples with final destinations.
The model was trained for 2 epochs, and then fine-tuned again with smaller learning rate only on final cities for 1 epoch.

\textbf{Decoding.}
The retrieval of items from the encoded sketch representation is done at the prediction stage. We retrieve scores for all items using the EMDE prediction procedure described in \S\ref{emde section}. Additionally, we post-process the output scores by multiplying by city popularity weights (calculated as the logarithm of the number of city occurrences as the final destination in a trip), thus boosting the scores of popular cities.
Finally, we pick 4 cities with the highest scores as the top predictions.


\textbf{Results.}
Our final score of Precision@4 metric on the  validation set is $0.601\%$.

\begin{table}[ht]
  \caption{Ablation study results.}
  \label{ablation results}
  \centering
  \small
  \setlength\tabcolsep{2.5pt}
  \begin{tabular}{c|ccccc}
\hline
Metric      & Basic & +Data & +Features & +Popularity & +Ensembling \\ \hline
Precision@4 & 0.552         & 0.573                  & 0.595           & 0.598               & 0.601      \\ 
Difference & - & +0.021 & +0.022 & +0.003 & +0.003  \\\hline
  \end{tabular}
\end{table}

\subsection{Ablation studies}
In order to understand the effect of crucial parts of the training process, we conduct additional experiments. To ensure a comparable number of parameters of all models, we adjusted the hidden size to have roughly the same total number of model parameters. 

\label{section:ablations}
The ablation results are summarized in Table \ref{ablation results}. In the \texttt{Basic} setting we train a pure EMDE model, which takes as input only the three concatenated city sketches (\textit{first city sketch}, \textit{prev city sketch}, \textit{all cities sketch}) and trains on final destinations only. This baseline model achieved  Precision@4 score of $0.552$. 
By including examples that are not final destination, and adding flag to the input that indicates if the examples is the last destination (\texttt{Data}),  we observed a $0.021$ increase of the precision score.
Concatenating continuous and categorical features to the input of neural network (\texttt{Features}) improves the precision score by $0.022$  compared to the model without features.
In addition to pure EMDE decoding we verify the impact of popularity boosting of final scores (\texttt{Popularity}). 
It increases Precision@4 score from 0.595 to 0.598. Furthermore, ensembling of 5 models (\texttt{Ensembling}) improves the precision by $0.003\%$

\begin{table}[!h]
  \caption{
  Performance of EMDE against of sequential models (without ensembling).}
  \label{baseline}
  \centering
  \small
  \begin{tabular}{c|ccc}
\hline
Metric      & EMDE & GRU + Cleora & GRU  \\ \hline
Precision@4 &  0.598 & 0.588  & 0.5786 \\ \hline
  \end{tabular}
\end{table}

In the next set of experiments, we compare EMDE against a sequential baseline model. The results are presented in Table \ref{baseline}. We train a GRU model \cite{cho2014properties} with hidden size of 1024 (selected empirically for best results), which takes as input either trainable city embeddings, or city embeddings learned by Cleora. All other input features are the same as in EMDE model. For fair comparison, we feed GRU outputs to the same three-layer feed-forward residual architecture as in EMDE. 
The application of sequential model decreases Precision@4 score from 0.598 to 0.588. In addition,
training GRU with randomly initialized trainable embeddings decreases Precision@4 score to 0.578.

\begin{table}[!h]
  \caption{Performance of our system (without ensembling) using various graph representation methods for computing input embeddings.}
  \label{ablation network}
  \centering
  \small
  \begin{tabular}{c|ccccc}
\hline
Metric      & Cleora  & Node2Vec & LINE & Word2Vec & GRU\\ \hline
Precision@4 & 0.5984  & 0.5956 & 0.5949 & 0.5907 & 0.5865\\ \hline
  \end{tabular}
  \label{ablation}
\end{table}

We also verify the impact of a graph embedding technique used to embed cities. We contrast Cleora with Word2Vec \cite{mikolov2013distributed} from the area of natural language processing, Node2Vec \cite{grover2016node2vec} which learns  a low-dimensional representations for nodes in a graph by optimizing a neighborhood preserving objective, and LINE \cite{tang2015line}, which  preserves the first-order node proximity and second-order node proximity separately, and then concatenates the two representations. 
In addition to network embedding techniques, we also compare to embeddings produced by a sequential model. 
We train a GRU model with trainable city embeddings to predict the final destination of a trip, and then use these learned embeddings to represent the cities as EMDE input.
The results are summarized in Table \ref{ablation network}. Cleora is a clear winner in terms of performance while also being significantly faster to train \cite{rychalska2021cleora}.

\begin{figure}[!h]%
\centering
\includegraphics[width=70mm]{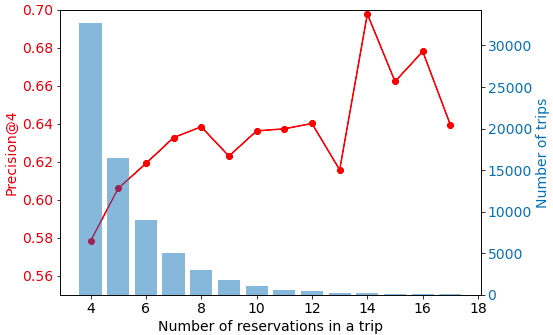}
\caption{Red plot presents the performance of the model in terms of number of reservations in a trip. The distribution of trip lengths is shown in the blue chart.}
\label{fig:distribution}
\end{figure}
\textbf{Error analysis.}
To discover possible sources of errors, we analyze predictions on different length of trips (Figure \ref{fig:distribution}). We notice that almost $50\%$ of reservations have only 4 bookings, and $97.6\%$ trips have less than 10 bookings. That taken into account, we observe that \textbf{the performance of our model increases with the number of reservations in a trip}. We hypothesize that it is easier for EMDE to capture dependencies in long sequences because in long trips the last part of the trip is often located in the same country (thus the cities are relatively close by), while in short sequences final locations can change abruptly.


In about $15\%$ validation examples, the last destination is the same as the first one. Our model achieves $0.902$ Precision@4 score on these examples, which leads to the conclusion that it learned to capture this particular phenomenon.

\begin{figure}%
\subfigure[Top 20 most popular countries]{
  \fbox{\includegraphics[width=38mm]{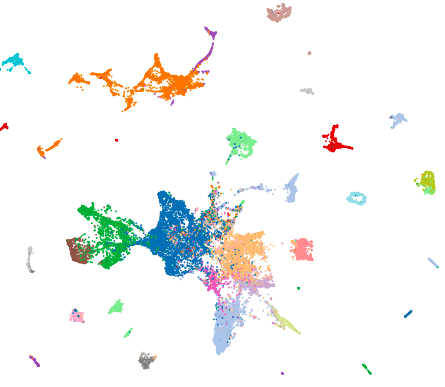}}
}
\subfigure[Top 60 to 80 most popular countries]{
  \fbox{\includegraphics[width=38mm]{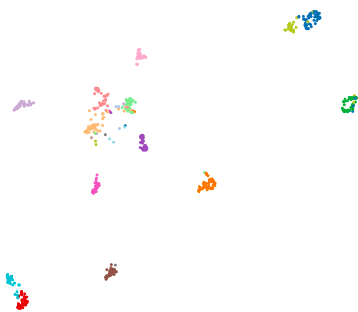}}%
}
\caption{2-D visualization of city embeddings learned by Cleora. Each color denotes a different hotel country.}
\label{fig:umap}
\end{figure}

\textbf{City embeddings visualization.}
Figure \ref{fig:umap} shows a visualization of high dimensional city embeddings mapped to 2-D space with UMAP \cite{mcinnes2018umap}. It can be seen that Cleora embeddings exhibit the awareness of geographic closeness of cities, which is well-represented irrespective of city popularity. As such, we show that the embeddings hold semantic knowledge which was not contained directly in the training graph (which was comprised of sequences of city IDs only, without any extra features).

\section{Summary}
In this paper we present our model which achieves 2nd place in Booking.com data challenge competition. 
We show that predicting the final city can be seen as a recommendation task. 
The system utilizes a graph embedding method to create multi-modal city vector representations, which are then encoded by EMDE into fixed-size \textit{sketch} structures. We show that accurate density estimation of sequences mapped to item sets can outperform inherently sequential methods.


\bibliographystyle{ACM-Reference-Format}
\bibliography{references.bib}

\end{document}